\documentclass[final]{cvpr}

\usepackage{times}
\usepackage{epsfig}
\usepackage{graphicx}
\usepackage{amsmath}
\usepackage{amssymb}
\usepackage{booktabs}
\def\B{\fontseries{b}\selectfont}
\usepackage{tabularx}
\usepackage{multirow}

\usepackage[pagebackref=true,breaklinks=true,colorlinks,bookmarks=false]{hyperref}

\begin{document}

\title{Skeletor: Skeletal Transformers for Robust Body-Pose Estimation}

\author{Tao Jiang, Necati Cihan Camg\"{o}z, Richard Bowden
\\
Centre for Vision, Speech and Signal Processing \\
University of Surrey, Guildford, UK \\
{\tt\small \{t.jiang, n.camgoz, r.bowden\}@surrey.ac.uk}
}

\maketitle
\newcommand{\SKEL}{\emph{Skeletor}}

\begin{abstract}
Predicting 3D human pose from a single monoscopic video can be highly challenging due to factors such as low resolution, motion blur and occlusion, in addition to the fundamental ambiguity in estimating 3D from 2D. Approaches that directly regress the 3D pose from independent images can be particularly susceptible to these factors and result in jitter, noise and/or inconsistencies in skeletal estimation. Much of which can be overcome if the temporal evolution of the scene and skeleton are taken into account. However, rather than tracking body parts and trying to temporally smooth them, we propose a novel transformer based network that can learn a distribution over both pose and motion in an unsupervised fashion. We call our approach \SKEL. \SKEL~overcomes inaccuracies in detection and corrects partial or entire skeleton corruption. \SKEL~uses strong priors learn from on 25 million frames to correct skeleton sequences smoothly and consistently. \SKEL~can achieve this as it implicitly learns the spatio-temporal context of human motion via a transformer based neural network. Extensive experiments show that \SKEL~achieves improved performance on 3D human pose estimation and further provides benefits for downstream tasks such as sign language translation.

\end{abstract}

\section{Introduction}

This paper introduces \SKEL, a deep neural network transformer model based on similar concepts to \emph{BERT} \cite{devlin2018bert}. \SKEL~is trained to correct noisy or erroneous 3D skeletal poses using implicit temporal continuity and a strong prior over body pose embedded within the transformer network. We demonstrate \SKEL's~application to correcting noisy 3D skeletal estimations that come from \emph{OpenPose} \cite{cao2018openpose} and the application of robust 3D pose estimation to downstream tasks such as sign language translation. Using a further state of the art transformer model, we show that the corrective power of \SKEL~provides significant benefits to downstream tasks. 

We train \SKEL~on a large corpus of human motion data, learning a probability distribution over the spatio-temporal skeletal motion. This distribution is learnt in an unsupervised manner, using a transformer network trained in a similar fashion to \emph{BERT} \cite{devlin2018bert}. However, whereas \emph{BERT} learns a language model over the discrete set of symbols that form words, \SKEL~learns to embed skeletal appearance in the context of human motion into the network. The output from \SKEL~is then a corrected 3D skeletal pose. It is important to note that although we demonstrate 3D skeletal correction, the approach is not limited to this domain. It would equally be possible to train the model on 2D skeletons or indeed any other regression problem where a large corpus of spatio-temporal data is available for training. 

Given a low quality video of a human gesturing (e.g. suffering from low resolution, motion blur and/or occlusion), our model can provide human motion estimation in moments of uncertainty by predicting the true 3D skeleton pose more precisely. This has significant benefits for fields such as gesture recognition, sign language translation or human motion analysis where poor quality data affects the performance of subsequent tasks. We train our model on 25 million frames to provide sufficient variety of human motion. Given the popularity of pose estimation, we believe \SKEL~will be a useful tool for many researchers. 

The contributions of this paper can be listed as: (1) We propose a novel transformer model that learns about the spatio-temporal manifold that represents 3D human skeleton pose. We call this model \SKEL~and demonstrate its application to correcting for missing limbs or whole skeletons within a sequence of human motion. It is able to correct noisy or erroneous joint positions; (2) \SKEL~is trained in an unsupervised manner, which allows it to be used on different datasets without human annotation; (3) \SKEL~can be used to further improve the accuracy of the downstream tasks. Experiments demonstrate that the application of \SKEL~improves performance for downstream tasks such as Sign Language Translation.

The remainder of the paper is structured as follows: In Section 2 we cover related work on human pose estimation and transformer networks and their origins in NLP. In Section 3 we describe the \SKEL~model in detail. We describe the training process and how to apply the model to 3D skeleton sequences in Section 4. In Section 5 we perform qualitative and quantitative experiments to demonstrate the effect of \SKEL~on 3D pose estimation. We then apply our model to the downstream task of sign language translation from skeletons, achieving improved performance and demonstrating the benefits of using \SKEL. Finally, we conclude the paper in Section 6, by discussing our findings and the future work. 

\section{Related work}

Human pose estimation has been an active area of computer vision since the 1980's. Early approaches were based on optimisation as either a 3D estimation problem, using model based approaches based on geometric primitives \cite{HOGG19835,Deutscher2005}, or part based 2D estimation \cite{Felzenszwalb03pictorialstructures}. The latter having the added complication that it suffers from self occlusion and perspective ambiguity. However, it was the introduction of Convolutional Pose Machines \cite{CPM}, building on the concept of pose machines proposed by Ramakrishna et. al. \cite{Ramakrishna2014} that provided a step change in performance. Since the introduction of Pose Machines, there has been significant momentum in this area, with numerous neural network based approaches to 2D estimation proposed, e.g. Stacked Hourglass Networks \cite{Newell} and HRNet \cite{HRNet}, to name but a few. The power of CNN approaches to pose estimation is not only to do with the learning approach, but also the sheer quantity of data they are trained on. The network learns extremely strong priors over the object pose.

When it comes to 3D estimation, there are two popular approaches: either taking a state of the art 2D regression approach and `lifting' this into 3D \cite{Martinez,Rochette2019} or attempting to regress the 3D directly from the image, often conditioned on initial 2D estimates \cite{xiang2019monocular,SMPL-X:2019}. Both approaches have their own advantages and disadvantages but in all cases, making use of the temporal cues to smooth out inaccuracies in single frame estimation improves results immensely \cite{xiang2019monocular}. As most approaches leverage the success of 2D estimation, failures in 2D detection directly relate to failures in 3D estimation. 

Strong models, which are trained on large variety of data, are key to overcoming visual ambiguities during estimation, but the model does not need to be an explicit 3D model, such as SMPL \cite{SMPL-X:2019}, as it can be learnt from data. In Natural Language Processing (NLP), strong language models are behind many recent success. A language model is a statistical model which provides the probability distribution of a fixed set of words. There have been several developments in NLP that are of relevance to this discussion. 

Firstly, embeddings: WORD2VEC was released in 2013 by Mikolov et. al \cite{word2vec}, which allowed a relatively simple neural network to learn the linguistic context of words from a large corpus of text. In addition to its popularity in NLP, it was quickly adopted by the vision community to solve problems such as image captioning \cite{7780872}. From an NLP perspective, further embeddings were developed such as GloVe (Global vectors for Word Representations) \cite{pennington2014glove} and FASTTEXT (Enriching word vectors with Subword Information)~\cite{bojanowski2017enriching}. 
ELMO \cite{peters2018deep} went beyond embeddings and used bidirectional LSTMs to learn a complex language model. However, ELMO was slow to train and suffered from the problem of long term dependencies. To overcome this, Bidirectional Encoder Representations from Transformers (\emph{BERT}) adopted the newer transformer-based architecture.

Transformer networks \cite{vaswani2017attention} are a relatively recent advancement that have achieved impressive results in many NLP and computer vision tasks, such as sign language recognition \cite{camgoz2020sign}, spotting \cite{varol2021read}, translation \cite{camgoz2020multi} and production \cite{saunders2020progressive}, object detection \cite{carion2020end}, scene segmentation \cite{fu2019dual} and video understanding \cite{sun2019videobert}. \emph{BERT} demonstrated how a pretrained language model, built from a large corpus of text, could be used to significantly outperform many state-of-the-art approaches across a broad range of down stream tasks such as question answering and language inference. A key observation, which is commonly accepted by the vision community, is that pretraining on a large corpus of data is key to network performance. We take the idea of \emph{BERT} and apply it in the context of human motion, learning a strong spatio-temporal prior from 25 million frames of video data. The learned prior improves the skeleton estimation accuracy, hence enhances the performance of downstream tasks relying on the estimated skeleton.

\section{\SKEL~Architecture}

\begin{figure}
\centering
\includegraphics[width=0.9\columnwidth]{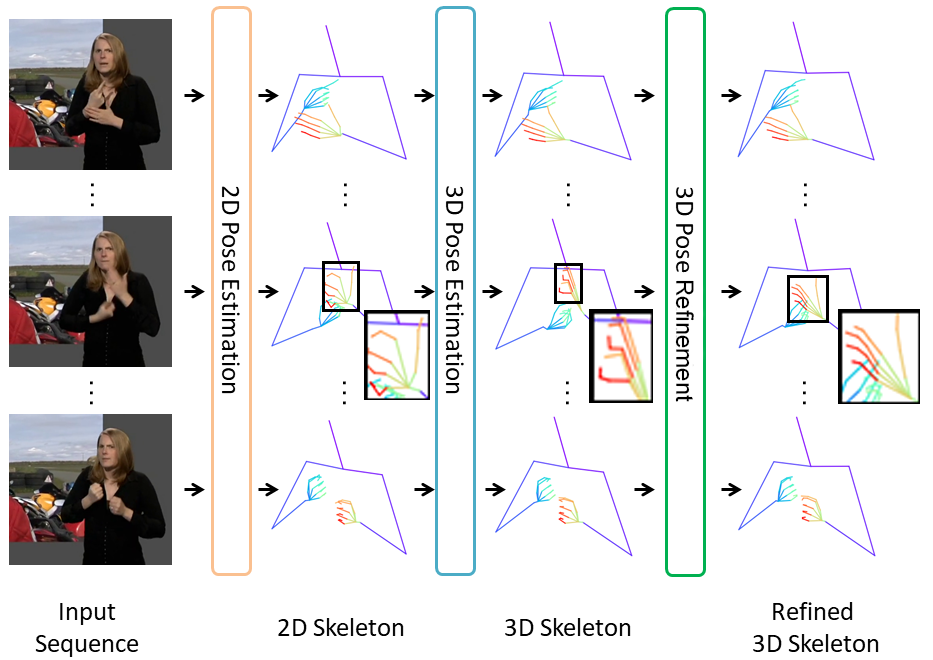}
\caption{Overview of our approach, which consists of three layers: 2D pose estimation, 3D pose estimation and refinement (\SKEL).}
\label{fig:2D->3D}
\end{figure}

In this section we introduce the architecture of \SKEL, a novel deep learning network that learns about the shape and motion of 3D skeletons from video in an unsupervised manner. Given an image sequence $V = (f_1, f_2, \cdots, f_T)$, our goal is to estimate an accurate 3D pose of the skeleton in each frame. We define 3D pose as $P = (P_1, P_2, \cdots, P_T)$, where $P_i = (J^i_1, J^i_2, \cdots, J^i_N)$ is the 3D skeleton in the $i^{th}$ frame and $J^i_k = (x^i_k, y^i_k, z^i_k)$ is the 3D position of joint $k$. An overview of our approach is seen in Figure \ref{fig:2D->3D}. First, \textbf{2D Pose Estimation} extracts a noisy and/or partially occluded 2D skeleton in image coordinates. This is then \emph{lifted} into 3D using either regression or inverse kinematics (IK) to produce a preliminary \textbf{3D Pose Estimation}, where errors in the 2D skeleton can effect the 3D estimate. \SKEL~ then provides \textbf{3D Pose Refinement} of the 3D skeleton.

The first step is to extract the initial \textbf{2D Pose Estimation} from a video sequence. There are many existing 2D pose estimation methods that could be employed \cite{cao2018openpose,cao2017realtime,fang2017rmpe}. Note that the output quality of 2D pose estimation is dependant upon both the technique used and the quality of the input images. Hence, any blur or occlusion in the original image sequence may cause failure in the estimated 2D pose. 

The next stage is to lift the 2D skeleton into 3D to obtain a \textbf{3D Pose Estimation}. Here we use state-of-the-art uplift methods \cite{zelinka2019nn,xiang2019monocular} to convert our initial 2D skeleton to 3D. The 3D pose accuracy is subject to the quality of the 2D pose, so it indirectly depends on the quality of the input images.

\begin{figure}
\centering
\includegraphics[width=0.85\columnwidth]{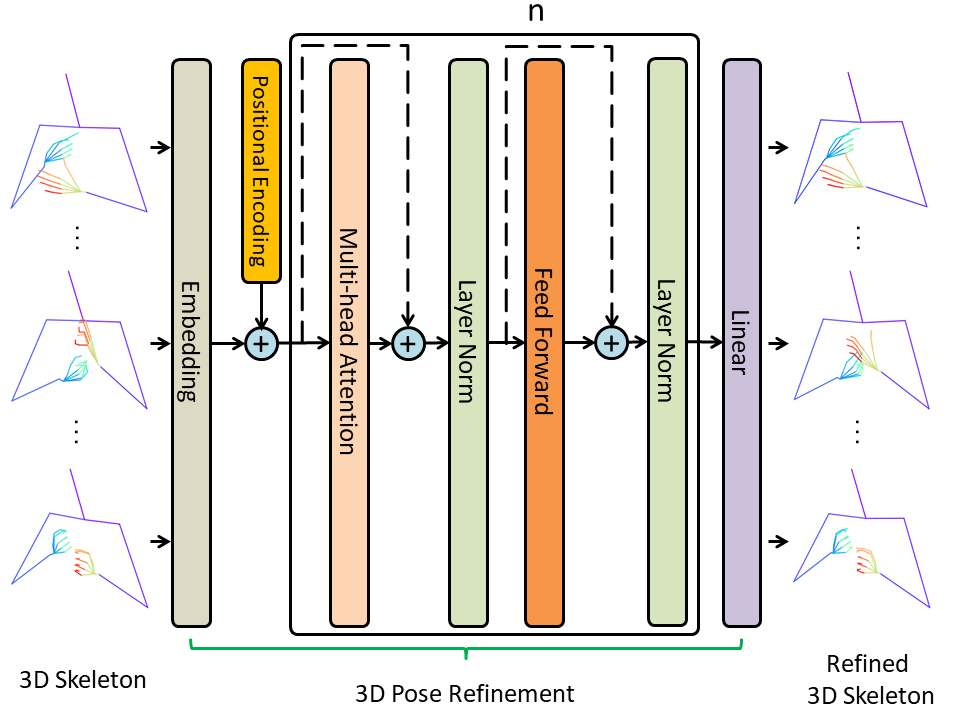}
\caption{The network architecture of the 3D Pose Refinement layer in \SKEL.}
\label{fig:network}
\end{figure}

The final stage is \textbf{3D Pose Refinement}, where we utilize the proposed BERT-inspired \SKEL~model. The quality of the 3D skeleton from uplift is dependant upon the quality of the video, motion blur or occlusions in the original images, which can lead to failures in the 3D skeleton sequence. \SKEL~solves this problem, by refining the inaccurate 3D skeleton based on its spatio-temporal context. Its architecture can be seen in Figure \ref{fig:network}.

Assuming we have a 3D skeleton sequence of $T$ frames, $ X = \{X_1, X_2, \cdots, X_T\}$, we first embed each skeleton into an embedding space with a dimensionality of $d_{model}$. Then positional encoding is added to the embedded skeleton to distinguish its order in the sequence. After that, the data goes through the encoder, which has $n$ identical layers. Each layer consists of two parts: a multi-head attention sub-layer and a position-wise feed-forward sub-layer. Finally, a linear network is employed to output the refined 3D skeleton sequence. In the rest of this section we introduce each component in detail.

\subsection{Skeletal Embedding}
Before feeding a skeleton sequence to the encoder, we embed the skeleton into a $d_{model}$ dimensional space. For poses with similar context, their embedded vectors should be closely located in the embedding space. The embedding is implemented by linear transformation, followed by ReLU activation and layer normalization \cite{ba2016layer} formulated as:
\begin{equation}
    \text{Embedding}(X_t) = \text{LayerNorm}(\text{ReLU}(\mathbf W_eX_t+\mathbf  b_e)),
\end{equation}
where $\mathbf W_e$ and $\mathbf b_e$ are learnable weights that project the skeleton vector to the embedding space. 

\subsubsection{Positional Encoding:} So that the model can account for the order of the skeletons in the sequence, we must inject information about the absolute or relative position of the skeletons in the sequence. To this end, we add a positional encoding vector to each input embedding. These vectors follow a specific pattern, allowing the model to determine the absolute position of each skeleton, or the relative distance between skeletons in the sequence. Here, we use sine and cosine functions of different frequencies to encode the skeleton positions:
\begin{equation}
\begin{aligned}
 \text{PE}(pos, 2i) = sin(pos/10000^{2i/d_{model}})\\
    \text{PE}(pos, 2i+1) = cos(pos/10000^{2i/d_{model}}),
\end{aligned}
\label{eq: logical 1}
\end{equation}
where $pos$ is the order of the frame in the sequence and $i$ is the dimension in the embedding space.

\subsection{Encoder} The encoder consists of a stack of $n=8$ identical layers. Each layer contains two sub-layers: a multi-head self-attention mechanism and position-wise fully connected feed-forward network. A residual connection \cite{he2016deep} is employed around each of the two sub-layers, followed by layer normalization. In order to facilitate the residual connections, all the sub-layers, as well as embedding layer, must produce outputs with the same dimension ($d_{model}$).

\subsubsection{Attention:} An attention function maps a query and a set of key-value pairs to an output, where the query and key are vectors of dimension $d_k$, while the value and output are vectors of dimension $d_v$. For each skeleton in the sequence, we create a query, key and value vector by multiplying the embedding of this skeleton by three matrices $W_Q, W_K, W_V$ that we learn during the training process. In practice, we pack all the queries on the whole sequence into a matrix $Q$. Similarly, the keys and values are packed together into matrices $K$ and $V$. Then the attention outputs are computed as:
\begin{equation}
    \text{Attention}(Q,K,V)=\text{softmax}(\frac{QK^T}{\sqrt{d_k}})V.
\end{equation}

In order to allow the model to jointly attend to information from different representation subspaces at different positions, we use Multi-Head Attention(MHA) to linearly project the queries, keys and values \emph{h} times with different, learned linear projections. This gives us \emph{h}, $d_v$-dimensional output vectors. We concatenate all the output vectors and project again to get the final MHA outputs.
\begin{equation}
\begin{aligned}
    \text{MultiHead}(Q,K,V)=\text{Concat}(\text{head}_1, \cdots, \text{head}_h)W^O,\\
    \text{where } \text{head}_i = \text{Attention}(QW_i^Q, KW_i^K, VW_i^V)
\end{aligned}
\end{equation}

where the projections are learned matrices $W_{i}^{Q} \in R^{d_{model} \times d_k}$, $W_i^K \in R^{d_{model}\times d_k}$, $W_i^V \in R^{d_{model}\times d_v}$ and $W^O \in R^{hd_v\times d_{model}}$.

\subsubsection{Position-wise Feed-Forward Networks:} The fully connected feed-forward network consists of two linear transformations with a ReLU activation in between. The function can be written as: 
\begin{equation}
    \text{FFN}(x) = \text{max}(0, x\mathbf W_1+\mathbf b_1)\mathbf W_2+\mathbf b_2
\end{equation}
The input and output have the same dimensionality $d_{model}$, while the inner-layer has dimensionality $d_{ff}$.

\section{Training and Evaluation}
In this section we describe the training and evaluation scheme for our models.

\subsection{Training}
We trained \SKEL~on a large corpus of TV broadcast footage, consisting of approximately 25 million high quality frames of human motion. We first run  OpenPose \cite{cao2018openpose} on the video footage to get the 2D skeletons and their confidence values. We then extract the 3D skeletons using Zelinka et al.'s 3D uplift process \cite{zelinka2020neural}, which can be summarized in the following steps. Firstly, the 3D coordinates of the skeleton's `head' joint are initialized as the coordinates of the target 2D skeleton. The length of each bone is then calculated by taking the average 2D length found in all frames of the current sequence. From the head joint, the 3D positions of all other joints' are computed recursively by optimizing the loss function, which contains three components; the MSE between the estimated 3D position's projection and 2D target position, the trajectory length from the last frame and the sum of the length of all the bones in the skeleton. There are 50 joints in total in each skeleton, that represent the upper-body movement.

For each sequence in a batch, some skeleton frames are partially or entirely corrupted by masking or by adding noise. The original skeleton and the associated corrupted skeleton are then fed into the model. We take the original skeleton as the ground truth due to the high quality of the dataset and use it to calculate the loss function. As loss function, we use the Mean Squared Error (MSE) between the prediction and the ground truth skeletons.

All parts of our network are trained with Xavier initialisation \cite{glorot2010understanding} and Adam optimization \cite{kingma2014adam} with default parameters and a learning rate of $10^{-5}$. Our code is based on Kreutzer et al.'s NMT toolkit, JoeyNMT \cite{JoeyNMT}, and implemented using PyTorch \cite{paszke2017automatic}. The whole training converged within 8 hours after 300,000 iterations. 

\subsection{Evaluation}
During training, we use a use a sliding window with fixed number of frames to create batches. However, for evaluation, the total number of frames in videos can differ and generally larger than the window size. To obtain final output of the $i^{th}$ frame in a video, we take the average of prediction at the same position in the $2r+1$ closest windows containing this frame, where $r$ is the averaging radius. In order to average the predictions, we need to pad $r + n_1 / 2$ frames both to the front and the rear of the video, where $n_1$ is the window size. To pad the video, we simply extend the first frame forwards and the last frame backwards.

\section{Results}
In this section, we first evaluate \SKEL~performance on corrupt and noisy skeletal sequences. We conduct experiments with masking and applying noise to both frames and joints, to showcase the skeleton correction abilities of \SKEL. We then apply \SKEL~on low quality videos to asses how much performance improvement can be provided. Finally, we evaluate the performance of \SKEL~in the context of a back translation task and make comparisons against the state-of-the-art.

\subsection{Masked Frames}
In our first experiment, we manually mask a percentage (5\%, 10\%, 15\%, 20\%, 25\%) of the frames in each sequence and use the unmasked skeletons as ground truth targets to train \SKEL. We use confidence values produced by 2D pose estimators (e.g. OpenPose) to pick frames with successful predictions (i.e. high confidence = good estimates). We consider the average joint confidence as each frame's confidence and choose the highest $p$ frames to mask. While validating, no matter what percentage we mask in training, we always validate on the same sequences with 15\% masked. As the mask is determined by the confidence, we always have the same mask, which guarantees a fair comparison among different experiment setups. 

We calculate the MSE on the development dataset at different iteration steps whilst training, reporting the minimum (min), average (ave) and maximum (max) MSE on the test data. The statistics in Table~\ref{tbl:exp1} show that using a 10\% masked model provides the best performance. In the following experiments, we continue to use this model if its not specified otherwise. 

\begin{table}[!h]
\centering
\caption{The left column is the frame masking percentage the model is trained with. The middle column is the MSE evaluated on the development dataset at different training steps. The right column is the minimum (min), average (ave) and maximum (max) MSE evaluated on test data. }
\label{tbl:exp1}
\resizebox{\linewidth}{!}{
\begin{tabular}{c|cccc|ccc}
 & \multicolumn{4}{c|}{Development (\# iterations)} &  \multicolumn{3}{c}{Test}\\ \hline
Frame mask & 50,000 & 100,000 & 200,000 & 300,000 &  min & ave & max \\\hline
 5\% & 3.086 & 2.124 & 1.195 & 1.103 & 0.147 & 1.833 & 9.628 \\
 10\% & {\B 2.613} & {\B 1.404} & {\B 0.642} & {\B 0.581} & {\B 0.090} & {\B 0.875} & {\B 8.421} \\
 15\% & 5.048 & 1.844 & 0.919 & 0.874 & 0.255 & 1.266 & 9.862 \\
 20\% & 5.518 & 2.196 & 1.034 & 0.856 & 0.264 & 1.250 & 9.459 \\
 25\% & 4.842 & 2.469 & 1.116 & 0.971 & 0.210 & 1.147 & 8.273 \\
\end{tabular}
}
\end{table}

\begin{figure}[]
\centering
\includegraphics[width=1.1\columnwidth]{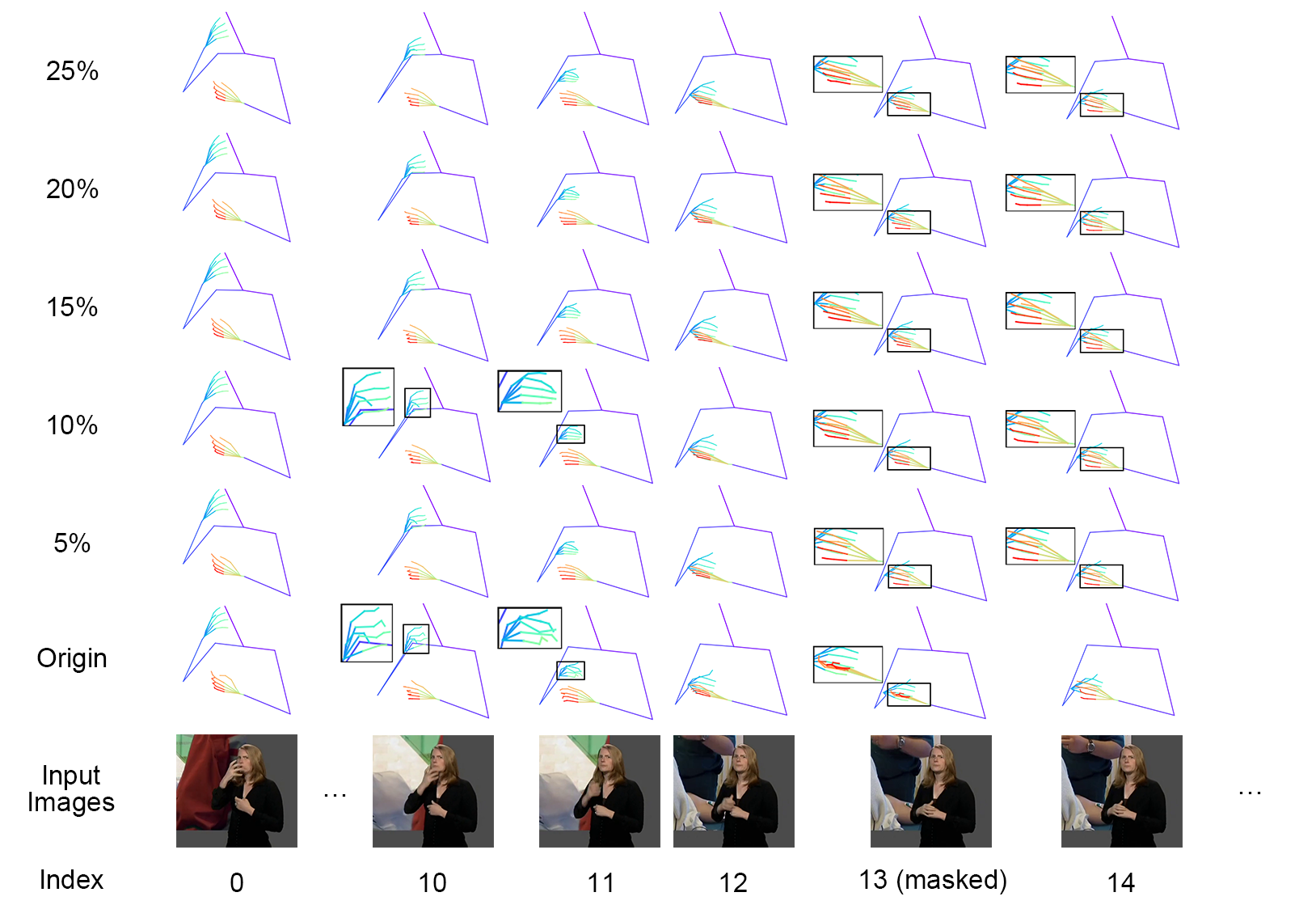}
\caption{The estimation results on test data from \SKEL~models trained with different frame mask percentages.}
\label{fig:exp1}
\end{figure}

For qualitative and intuitive comparison, we also plot the prediction results for different models in Figure \ref{fig:exp1}. From the figure, we can see that \SKEL~not only predicts masked frames (frame 13) but is also tolerant to noisy data in training (frame 10, 11). For prediction on masked frame 13, a 5\% masked model simply copies from the preceding frame. However, as masking increases to 10\% and beyond, the model starts predicting the missing skeleton according to the motion context that has been previously learnt.

\subsection{Masked Joints}
In our next experiment, we mask out the joints instead of the frames, evaluating the ability of \SKEL~to replace missing joints. We mask a certain percentage (5\%, 10\%, 15\%, 20\%, 25\%) of the highest confidence joints in the sequences rather than the whole skeleton of a frame. While testing, a fixed 15\% of random joints are masked, choosing the same random seed as previous so that we get a consistent mask for fair comparison. Evaluation of the different joint masking percentages can be found in Table \ref{tbl:exp2}, showing that the model with 10\% joint masking achieves the best results. 

\begin{table}[!h]
\centering
\caption{The MSE results on the development and test datasets for models with different joint mask percentages.}\label{tbl:exp2}
\resizebox{\linewidth}{!}{
\begin{tabular}{c|cccc|ccc}
 & \multicolumn{4}{c|}{Development (\# iterations)} &  \multicolumn{3}{c}{Test}\\ \hline
 Joint mask & 50,000 & 100,000 & 200,000 & 300,000 &  min & ave & max \\\hline
 5\% & \textbf{{17.532}} & 15.146 & 13.655 & 13.645 & 2.019 & 4.543 & 12.484 \\
 10\% & 18.192 & \textbf{{14.493}} & \textbf{{13.585}} & \textbf{{13.521}} & \textbf{{1.869}} & \textbf{{3.555}} & \textbf{{9.590}} \\
 15\% & 20.759 & 15.594 & 14.585 & 14.640 & 1.979 & 4.181 & 15.478 \\
 20\% & 23.323 & 18.590 & 17.881 & 17.914 & 2.606 & 6.390 & 42.758 \\
 25\% & 23.453 & 17.816 & 16.452 & 16.418 & 2.314 & 5.031 & 22.171 \\
\end{tabular}
}
\end{table}

\begin{figure*}[]
\centering
\includegraphics[width=0.73\textwidth]{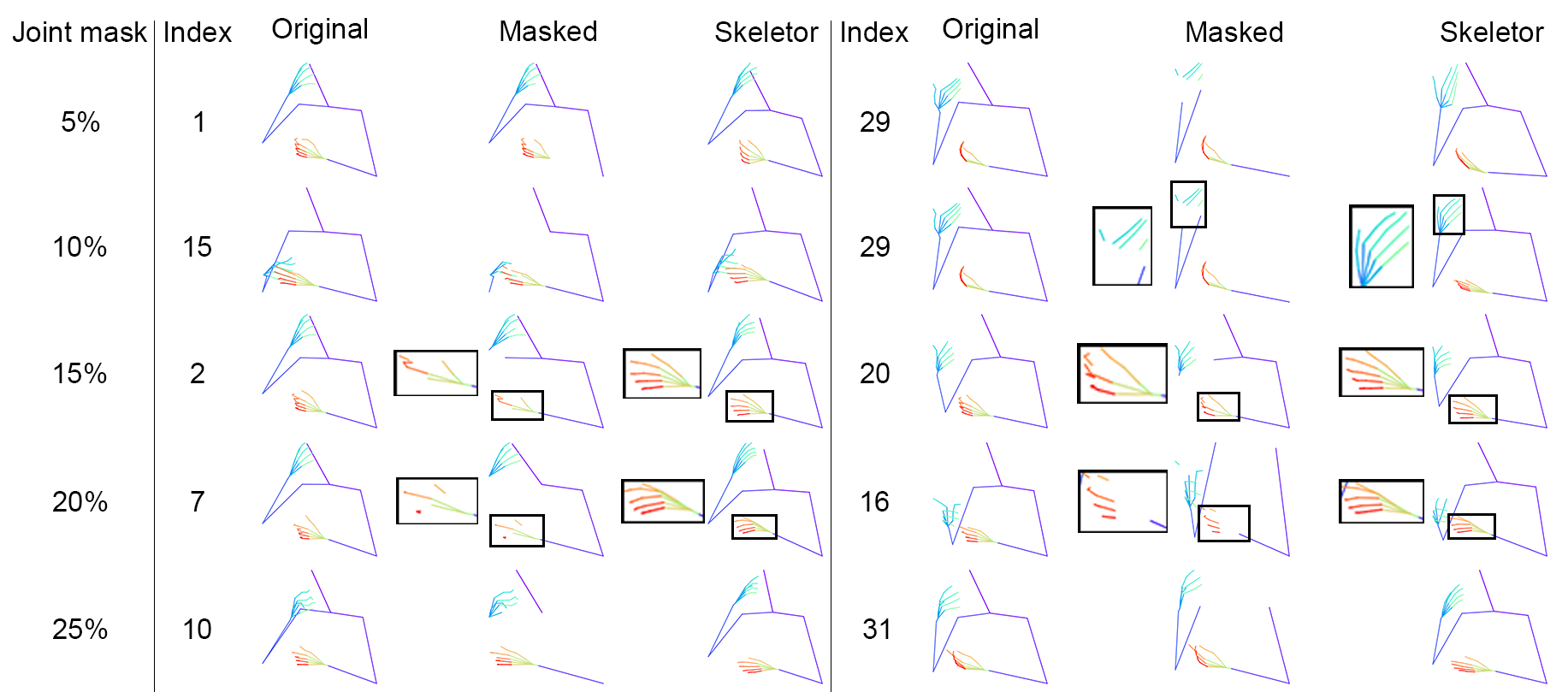}
\caption{The joint masked skeletons with different masking percentages corrected by \SKEL. The first column is the joint masking percentages. Each column in 2-3 includes frame indices, original skeletons, joint masked skeletons and predictions from \SKEL.}
\label{fig:exp2}
\end{figure*}

We visualize the masked skeletons and \SKEL~predictions in Figure \ref{fig:exp2}. From the figure, we can see that \SKEL~can complete missing joints without exception. Although sometimes the gesture cannot be recovered accurately (e.g. the left hand in frame 31 on 25\% of the joint masked model), the whole-body pose and the general hand shape given by \SKEL~are close to the original. This demonstrates that \SKEL~can not only learn the context of the motion but also learn the context of the joint movement.

\subsection{Noisy Frames}

To demonstrate that \SKEL~can correct erroneous skeleton sequences, we manually corrupt the high-quality data by adding noise to the 3D skeleton during testing. The noise strength is determined by the parameter $s$. We add noise to the $i^{th}$ joint's position $n_i\in[-s*limb_i,s*limb_i]$, where $limb_i$ is the length of the limb whose end is joint $i$. $n_i$ is uniformly and randomly chosen between $-s*limb_i$ and $s*limb_i$. We corrupt 15\% of the highest confidence frames in the sequence with different noise strengths (0.1, 0.3, 0.5, 0.7, 0.9) to see how much noise \SKEL~can tolerate. 

\begin{table}
\centering
\caption{\SKEL~can be used to correct the skeleton with noise applied to the whole skeleton.}\label{tbl:exp3}
\resizebox{0.7\linewidth}{!}{
\begin{tabular}{c|ccc}
Frame-level & \multicolumn{3}{c}{MSE on Test set}\\ \hline
Noise Strength ($s$) &  min & ave & max \\\hline
 0.1 & 0.059 & 0.117 & 1.408 \\
 0.3 & 0.154 & 0.215 & 1.488 \\
 0.5 & 0.345 & 0.409 & 1.691 \\
 0.7 & 0.585 & 0.689 & 1.983 \\
 0.9 & 0.841 & 1.100 & 2.331 \\
\end{tabular}
}
\end{table}

\begin{figure*}[t]
\centering
\includegraphics[width=.85\textwidth]{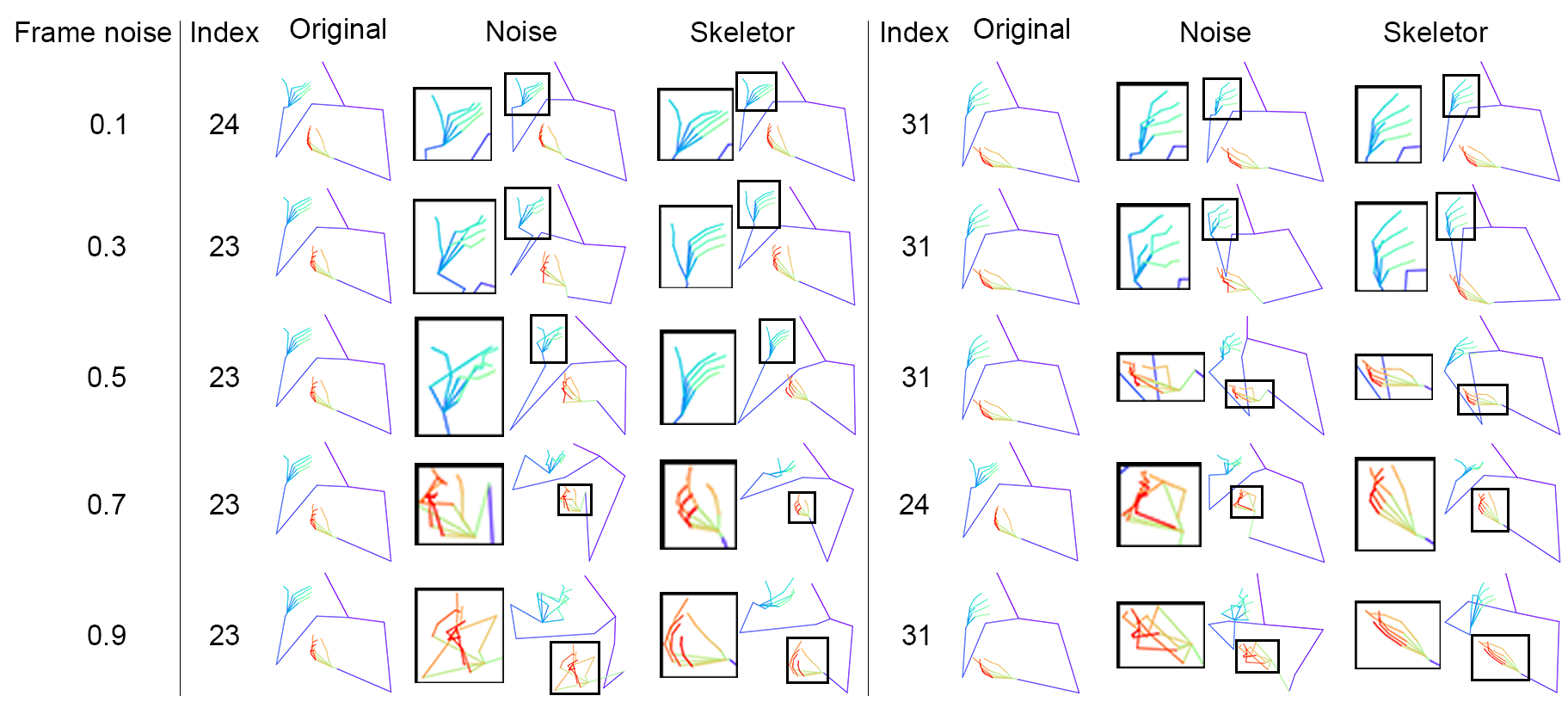}
\caption{The noisy skeletons with different noise strengths corrected by \SKEL. The first column is the frame indices and the original skeleton. Each column in 2-6 has two parts. The left part is the noisy skeletons corrupted with noise (strength noted in parentheses), while the right part are the results recovered by \SKEL. }
\label{fig:exp3}
\end{figure*}

The estimated errors can be found in Table \ref{tbl:exp3}. Obviously, with less noise, \SKEL~achieves better results. For intuitive observation, we also draw the noisy skeleton and \SKEL~prediction in Figure \ref{fig:exp3}. From the figure, we can see that \SKEL~can recover the original gesture well with a noise strength below 0.3. However, above 0.5, \SKEL~begins to struggle. Although it cannot perfectly estimate the whole-body pose under the influence of heavy noise, it can still give recognizable results based on the whole-body pose (especially for noise 0.9).

\subsection{Noisy Joints}

\begin{figure*}
\centering
\includegraphics[width=.85\textwidth]{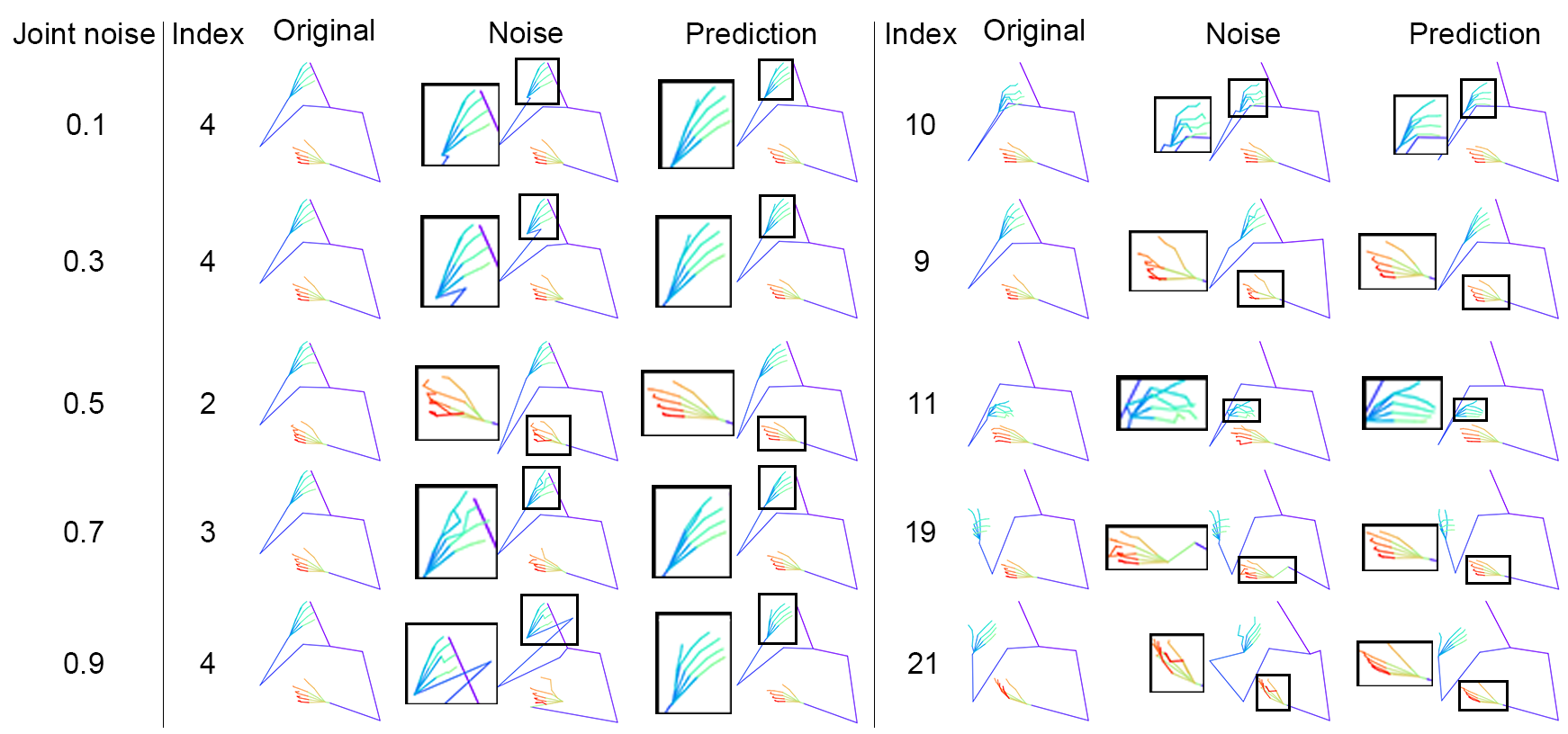}
\caption{\SKEL~can be used to correct different levels of joint inaccuracies in the skeleton.}
\label{fig:exp4}
\end{figure*}

In order to demonstrate that \SKEL~can correct erroneous joints in the sequence, during testing we corrupt 15\% of the highest confidence joints instead of the whole skeleton. We use different noise strengths (0.1, 0.3, 0.5, 0.7, 0.9). Quantitative results can be found in Table~\ref{tbl:exp4}, with \SKEL~performing better with less noise. However, in Figure~\ref{fig:exp4}, we can see that \SKEL~corrects the joint position and provides a sensible prediction. Even with a large noise strength applied to the joint (noise 0.9 in the 4-th frame), \SKEL~can produce results comparable with much smaller amounts of noise.

\begin{table}
\centering
\caption{\SKEL~can be used to correct the skeleton with different levels of noise applied to the joints.}\label{tbl:exp4}
\resizebox{0.7\linewidth}{!}{
\begin{tabular}{c|ccc}
Joint-level& \multicolumn{3}{c}{MSE on Test set}\\ \hline
Noise Strength ($s$) &  min & ave & max \\\hline
 0.1 & 0.225 & 0.452 & 4.573 \\
 0.3 & 0.272 & 0.508 & 4.786 \\
 0.5 & 0.364 & 0.616  & 5.126 \\
 0.7 & 0.496 & 0.770 & 5.209 \\
 0.9 & 0.656 & 0.965 & 5.483 \\
\end{tabular}
}
\end{table}

\begin{figure*}[!h]
\centering
\includegraphics[width=.85\textwidth]{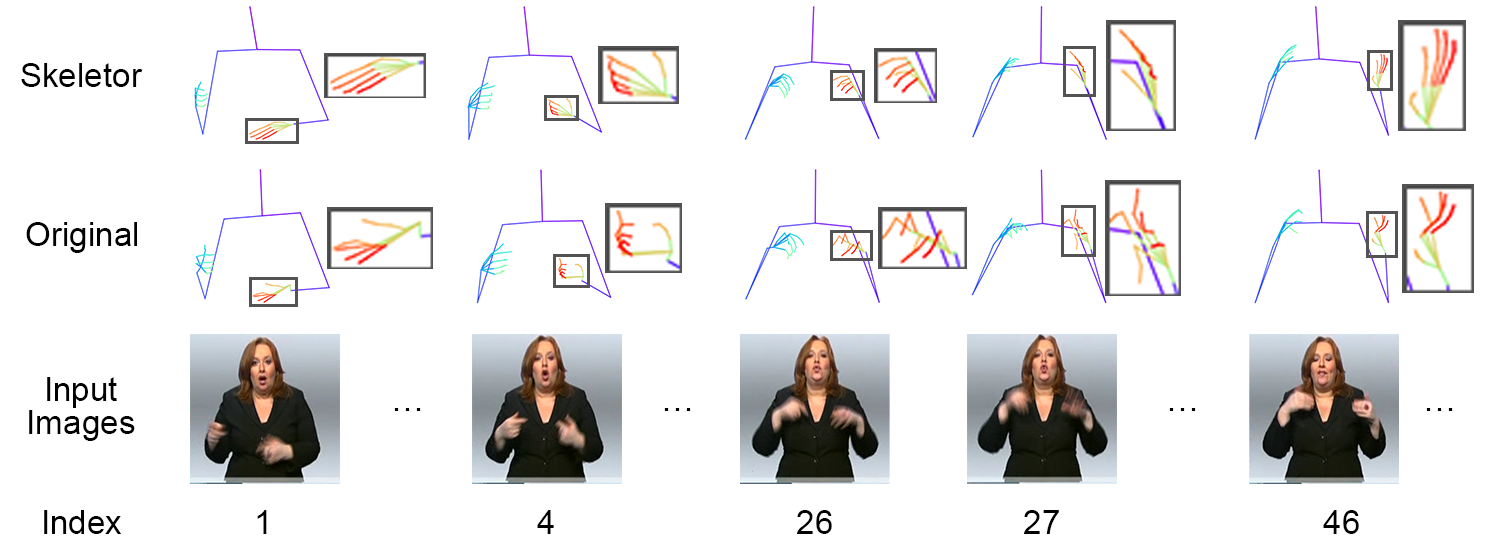}
\caption{\SKEL~can be applied to datasets with low image quality to improve the estimation accuracy}
\label{fig:exp8}
\end{figure*}

\begin{table*}
\centering
\caption{Comparison with the Sign2Text baseline sign language translation model, using the back translation evaluation.}\label{tbl:backtrans}
\resizebox{\textwidth}{!}{
\begin{tabular}{l|ccccc|ccccc}
& \multicolumn{5} {c|}{DEV SET} & \multicolumn{5}{c}{TEST SET} \\ \hline
& BLEU-4 & BLEU-3 & BLEU-2 & BLEU-1 & ROUGE & BLEU-4 & BLEU-3 & BLEU-2 & BLEU-1 & ROUGE \\ \hline
Sign2Text \cite{camgoz2018neural} &  9.94      & 13.16       & 19.11       & 31.87       & 31.80      & 9.58       & 12.83       & 19.03       & {\B 32.24}       & 31.80  \\
Raw 3D Skeleton Estimates & 9.74   & 12.75  & 18.38  & 31.43  & 31.82 & 8.85   & 11.62  & 16.94  & 30.22  & 29.89 \\
\SKEL~& {\B 10.91}  & {\B 14.01}  & {\B 19.53}  & {\B 31.97}  & {\B 32.66} & {\B 10.35}  & {\B 13.49}  & {\B 19.11}  & 31.86  & {\B 31.80} \\
\end{tabular}
}
\end{table*}

\subsection{Low Quality Video}

For all previous experiments, the data we have trained and tested upon has been of high quality, with synthetic noise or masking applied. After training our model, we want to see to what extent \SKEL~can improve the nature of low quality data. We apply \SKEL~to the publicly available PHOENIX14T dataset \cite{camgoz2018neural}, which is recorded from broadcast footage and suffers from both motion blur and interlacing artifacts. As there is no 3D ground truth for this dataset, we can only evaluate from a qualitative perspective. 

In Figure~\ref{fig:exp8} we see that when motion blur occurs, it is difficult for both the human eye or machine to understand the sign from a single image. This can lead to prediction failure (see the left hand in the row labelled Original), which would affect the performance of downstream tasks training on this dataset. However, \SKEL~learns the context and motion pattern from the whole sequence rather than independent images. It can use the context of the sequence and visual cues from when the hands are not blurred, to infer the pose and motion throughout the entire sequence, \SKEL~predicts the missed skeleton or corrects the erroneous pose estimation from the spatio-temporal context it has learned (see the left hand in the \SKEL~row).

\subsection{Back translation}

To evaluate the effect of \SKEL~on the performance of downstream tasks, we trained a separate transformer based Sign Language Translation network on the PHOENIX14T dataset. Our network was built using 2 layers, which have hidden layer size of 64 and 256 feed forward units each. We utilized Adam optimizer \cite{kingma2014adam} with a learning rate of $10^{-3}$ and a batch size of 32. We also employed 0.1 dropout and $10^{-3}$ weight decay. As defined by the dataset evaluation protocols \cite{camgoz2018neural}, we measure the translation performance of our model using BLEU \cite{papineni2002bleu} and ROUGE \cite{lin2004rouge} scores, which are commonly used in the field of machine translation.

As can be seen in Table~\ref{tbl:backtrans}, \SKEL~improves the translation performance by a relative 12\%/17\% BLEU-4 scores on dev/test sets respectively. Furthermore, our model was also able to surpass the Sign2Text baseline, which goes directly from video to spoken language text.

\section{Conclusion}
In this paper, we proposed a novel transformer based network called \SKEL, which can learn the spatio-temporal context of human motion at both the skeletal animation and joint level. Within this context, it can not only predict the missing joints and skeletons in the sequence, but also correct noisy skeletons or joint inaccuracies. Unsupervised training makes it possible to learn from and leverage significant benefits from huge corpora of data. 

Our experiments showed that \SKEL~improves the accuracy of pose estimation, especially where we have low quality video. Applying masking and noise augmentation at both frame- and joint-level provides a boost in both quantitative and qualitative performance. Our corrected skeletons achieve improved results on sign language translation, demonstrating that \SKEL~can have significant benefits to down stream tasks. 

In the future, we will investigate the addition of an adversarial discriminator to further increase the realism of skeletal prediction. Another future direction is to add sequence ordering mechanism as adopted by BERT to detect discontinuities in the sequence.

\section*{Acknowledgements}
This work received funding from the SNSF Sinergia project `SMILE II' (CRSII5\_193686), the European Union's Horizon2020 research and innovation programme under grant agreement no. 101016982 `EASIER' and the EPSRC project `ExTOL' (EP/R03298X/1). This work reflects only the authors view and the Commission is not responsible for any use that may be made of the information it contains.

{\small
\bibliographystyle{ieee_fullname}
\bibliography{egbib}
}

\end{document}